\documentclass{article}
\usepackage[utf8]{inputenc}
\usepackage{authblk}
\usepackage{setspace}
\usepackage[margin=1.25in]{geometry}
\usepackage{graphicx}
\graphicspath{ {./figures/} }
\usepackage{subcaption}
\usepackage{amsmath}

\usepackage[style=nejm, 
citestyle=numeric-comp,
sorting=none]{biblatex}
\title{An Ergonomic, Customizable Soft Robotic Glove toward Personalized Hand Rehabilitation}

\author[1*]{Rui~Chen}
\author[2]{Firman~Isma~Serdana
}
\author[1]{Domenico~Chiaradia}
\author[1,3]{Xianlong~Mai
}
\author[2,4]{Elena~Losanno
}
\author[5]{Gabriele~Righi
}
\author[5]{Claudia~De~Santis
}
\author[1]{Federica~Serra
}
\author[6]{Vincent~Mendez
}
\author[1]{Cristian~Camardella
}
\author[1]{Daniele~Leonardis}
\author[5]{Giulio~Del~Popolo}
\author[2,4,6]{Silvestro~Micera}
\author[1]{Antonio~Frisoli}

\affil[1]{Institute of Mechanical Intelligence and Department of Excellence in Robotics and AI, Scuola Superiore Sant'Anna (SSSA), 56127, Pisa, Italy
}
\affil[2]{The Biorobotics Inst. and Dept. of Excellence in Robotics and AI, Scuola Superiore Sant’Anna, Pisa, Italy
}
\affil[3]{CAS Key Laboratory of Mechanical Behavior and Design of Materials, Institute of Humanoid Robots, School of Engineering Sciences, University of Science and Technology of China, Hefei, China
}
\affil[4]{Modular Implantable Neuroprostheses (MINE) Laboratory, Università Vita-Salute San Raffaele \& Sant’Anna School of Advanced Studies, Milan, Italy
}
\affil[5]{ Careggi University Hospital, 50134 Florence, Italy
}
\affil[6]{Bertarelli Fndn. Chair in Translational Neuroengineering, Neuro X Institute, École Polytechnique Fédérale de Lausanne (EPFL), Laussane, Switzerland
}
\affil[*]{Corresponding author. Email: rui.chen@santannapisa.it}

\date{}

\begin{document}

\maketitle
%%%%%%%%%%%%%%%%%%%%%%%%%%%%%%%%%%%%%%%%%%%%%%%%%%%%%%%%%%%%%%%%%%%%%%%%%%%%%%%%%%%%%%%%%%%%%%%%%%%%%%%%%%%%%%%%%%%%%%%%%%%%%%%%%%%%%%%%%%%%

\begin{abstract}
Hand impairment following neurological disorders substantially limits independence in activities of daily living, motivating the development of effective assistive and rehabilitation strategies. Soft robotic gloves have attracted growing interest in this context, yet persistent challenges in customization, ergonomic fit, and user comfort constrain their clinical utility. Here, we present an ergonomic, customizable fabric-based soft robotic glove whose actuators can be tailored to individual finger-joint geometry. The glove comprises five dual-action actuators supporting finger flexion and extension, together with a dedicated thumb abduction actuator. Leveraging computer numerical control heat sealing technology, we fabricated symmetrical-chamber actuators that adopt a concave outer surface upon inflation, thereby increasing finger contact area and improving comfort. Characterization confirmed joint moment and grasping force sufficient for ADL-relevant tasks. In ten healthy subjects, active assistance significantly reduced forearm muscle activity during manipulation, and a pilot study in three individuals with cervical spinal cord injury showed more natural grasp patterns and reduced reliance on tenodesis grasp.
\end{abstract}


\section*{Introduction}

The human hand is central to a broad range of activities of daily living (ADLs)~\cite{kottink10therapeutic_effect_glove}, and neurological disorders such as stroke~\cite{feigin2022_Stroke_Sheet,park2024robot_treatment_for_stroke}, cervical spinal cord injury (SCI)~\cite{hu2023SCI}, and muscular dystrophy~\cite{roberts2023muscular_dystrophy} can substantially impair it, often reducing ADL independence~\cite{Seth2024_Disease_stroke,park2024robot_treatment_for_stroke,Stella2006_Task-Specific}. Cervical SCI is illustrative: affected individuals frequently rely on tenodesis grasp, a passive strategy in which wrist extension tightens the finger flexors to produce a weak closing force~\cite{pripotnev2023tenodesis_grasp,jung2018Tenodesis_Grasp} that is often insufficient for reliable manipulation~\cite{cappello2018_Assisting_silicone_glove,Correia2020_Improving_Grasp}. Restoring hand function in such populations requires not only active finger flexion but also powered extension for intentional object release~\cite{lim2023_Bidirectional_Fabric,kim2022bioinspired—_cable_driving}, together with thumb abduction, a motion critical for prehensile grasping~\cite{nanayakkara2017role_thumb,ingram2008statistics_hand_movement_thumb,xie_2024_abduction,DEKRAKER2009_Thumb_Abduction}.

Robotic rehabilitation has emerged as a promising paradigm, offering controlled and repeatable assistance that supports functional recovery~\cite{Gassert2018_Rehabilitation_Robot,park2024robot_treatment_for_stroke}. Among robotic rehabilitation devices, soft robotic gloves have attracted considerable interest~\cite{tiboni2022_soft_glove_review}. By employing compliant materials and flexible actuation mechanisms, these devices can approximate natural hand motion while reducing the mechanical constraints associated with rigid exoskeletons~\cite{maceira2019_Wearable_Sensors,Correia2020_Improving_Grasp,Pan2015_silicone_2015_first}. A range of actuation technologies has been integrated into soft glove designs, including cable-driven systems~\cite{bagneschi_2023_cable_soft_hand,2016_polymer_cable_Driven,Alicea_2021_cable_synergy_glove,kim2022bioinspired—_cable_driving}, shape memory alloys~\cite{Lai2023SMA,lee2024_SMA}, elastomeric pneumatic actuators~\cite{Hong2016_elastomeric_silicon_glove,Pol2015_EMG_control_glove,cappello2018_Assisting_silicone_glove,Suulker_comparion_silicone_fabric}, and fabric-based pneumatic actuators~\cite{Yap2016fabric-regulated,Ge2020_Design_Fabric,lim2023_Bidirectional_Fabric,Lai2023_Honey_Comb_glove,Feng2021_Asymmetric_Fabric,yap2017_fully_fabric_bidirectional,lim2023_Bidirectional_Fabric,yap2017design_low_cost_fabric}. Despite this breadth of approaches, persistent challenges in customization, ergonomic fit, and user comfort continue to limit the clinical utility of existing designs~\cite{Low2015Customizable_Soft}.

Per-user fit and ergonomic conformity are two challenges that bear directly on actuator design. First, hand size and joint geometry vary considerably across individuals, so an actuator built to a generic geometry tends to misalign with a given user's joint axes, introducing unintended joint torques that reduce force-transmission efficiency and comfort~\cite{Poly2015_Fiber_actuator}; effective assistance therefore requires a fit tailored to each user, yet rigid or modular designs achieve size adaptability at the cost of added weight and complexity~\cite{Low2015Customizable_Soft,Kokubu2024_Personalizd_glove2,kokubu2024_personalized_glove,yun2017exo-glove_pm}, and per-individual geometric tailoring remains uncommon among soft and lightweight designs. Second, single-chamber pneumatic actuators inflate into a convex shape; because the finger dorsum is itself convex, the two convex surfaces meet over only a small area, limiting contact and comfort~\cite{Hong2016_elastomeric_silicon_glove,cappello2018_Assisting_silicone_glove,Feng2021_Asymmetric_Fabric}. Pleated or folded fabric structures can improve joint alignment but still present a convex contact profile~\cite{lee2024_ergonomic,liu2023finger_joint_aligned_folding}. No existing soft rehabilitation glove combines ergonomic concave contact and per-user geometric customizability with dual-action assistance and thumb abduction in a single lightweight fabric device (Supplementary Tables~4 and~5).

To address these challenges, we present an ergonomic, customizable fabric-based soft robotic glove for hand rehabilitation. Each actuator is formed from two symmetric chambers and a bottom constraint layer, shaped by CNC heat sealing, and this single design provides two properties at once: the symmetric chambers inflate into a concave contact surface that conforms to the convex finger dorsum, increasing contact area and comfort; and the CNC-defined chamber profile sets the inflated bending angle, allowing each actuator to be tailored to a wearer's individual joint angles. These actuators are assembled into a single lightweight (90~g) all-fabric glove providing active flexion, extension, and thumb abduction across all five digits. We characterized actuator moment and force output, evaluated whole-glove grasping performance, conducted a preliminary surface electromyography (sEMG) study with healthy subjects to quantify muscle-activity reduction during assisted manipulation, and carried out a pilot feasibility study in individuals with cervical SCI, providing validation across the device, healthy-subject, and clinical levels.

\section*{Results}

\subsection*{Glove Design}

\begin{figure}[htbp]
    \centering
    \includegraphics[width=\columnwidth]{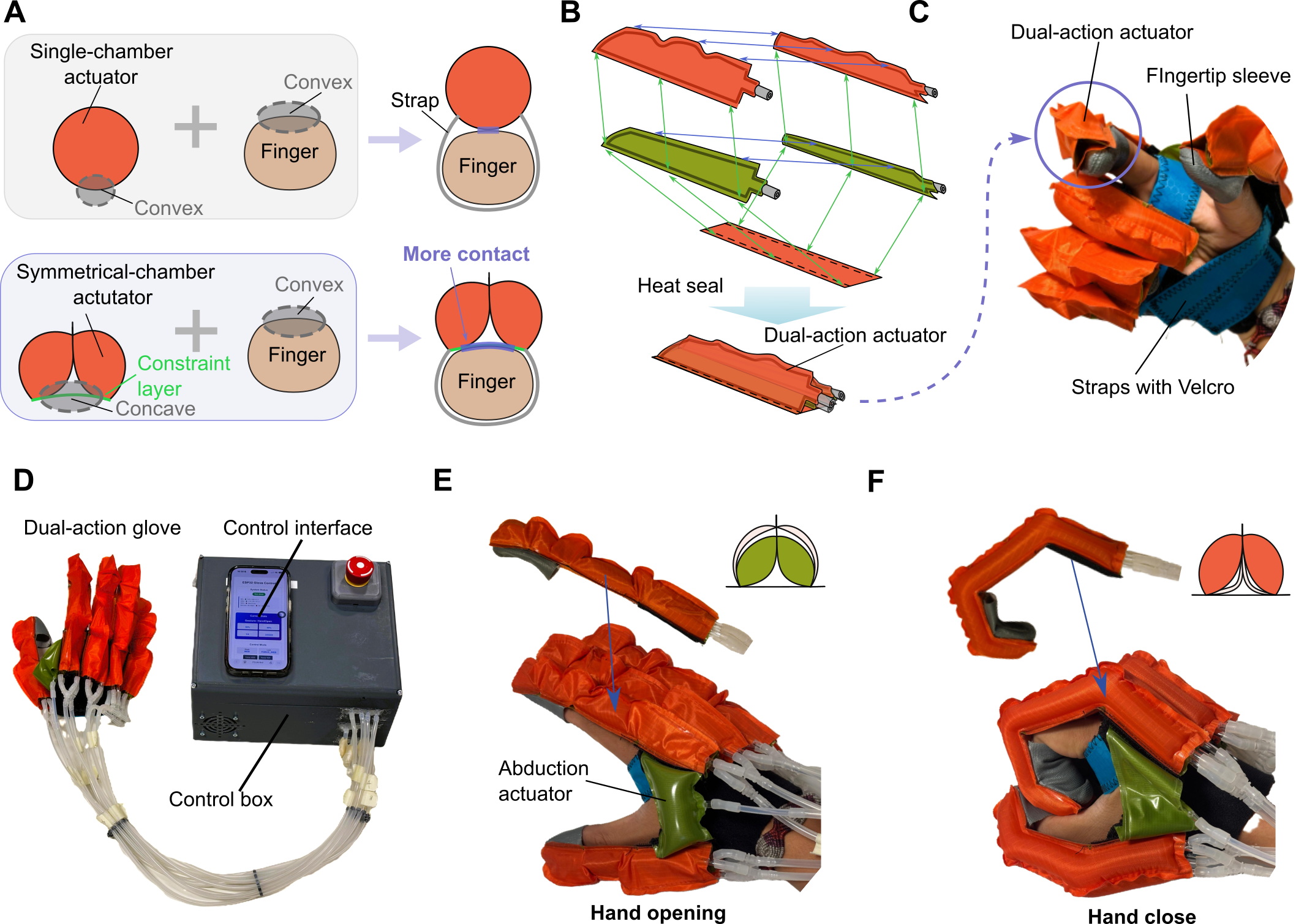}
    \caption{\textbf{Glove system overview.}
    \textbf{(A)} Comparison of finger contact between single-chamber and symmetrical-chamber actuators upon inflation.
    \textbf{(B)} and \textbf{(C)} Design of the dual-action actuator and soft glove.
    \textbf{(D)} Complete system comprising the control box, dual-action soft glove, and smartphone web interface.
    \textbf{(E)} and \textbf{(F)} Glove in fully extended and fully flexed configurations, respectively.}
    \label{Fig-M-Conceptual}
\end{figure}

Each actuator is built from two symmetric chambers and a bottom constraint layer (as baffle) that restricts outward expansion, with the chambers shaped by CNC heat sealing. A conventional single-chamber actuator inflates into a convex shape that, meeting the convex finger dorsum, contacts the finger over only a small area. By contrast, the two symmetric chambers cause the actuator to adopt a slightly concave outer surface upon inflation, which conforms to the finger dorsum and increases the contact area (Fig.~\ref{Fig-M-Conceptual}A), distributing contact pressure more evenly and improving wearing comfort.

The same two chambers are profiled by CNC heat sealing; because the deflated profile geometry maps near-linearly to the inflated bending angle~\cite{chen2026_GPAs}, the required profile can be computed for any target joint angle, so each actuator is also customizable to a wearer's hand. Whereas our prior symmetrical-chamber glove provided flexion assistance only~\cite{Rui_2025_customized_glove}, here each finger actuator pairs a flexion and an extension chamber set assembled by CNC heat sealing (Fig.~\ref{Fig-M-Conceptual}B), so that selective inflation drives the finger toward flexion or extension. The glove consists of a fabric base, five dual-action finger actuators with fingertip sleeves, one thumb abduction actuator, and three Velcro-secured straps (Fig.~\ref{Fig-M-Conceptual}C). All actuators are mounted to the fabric base via Velcro, enabling straightforward adjustment and rapid replacement as needed. The open strap-and-sleeve configuration facilitates donning and doffing, which is particularly important for patients with limited hand function. The assembled glove weighs only 90~g and is entirely fabric-based, providing a lightweight and compliant interface well suited to prolonged rehabilitation use.

A custom control box provides independent inflation control of the flexion and extension chambers of each finger via a smartphone web interface (Fig.~\ref{Fig-M-Conceptual}D). This per-finger capability is intended to support task-oriented actuation strategies~\cite{li2024Task-oriented,lee2024Task-oriented}, although the present study employed only whole-hand flexion and extension commands. Full hand opening and full hand closure are illustrated in Fig.~\ref{Fig-M-Conceptual}E and F.

\subsection*{Glove Fabrication}

Advances in CNC heat sealing technology for fabrics~\cite{nguyen2020design_CNC,gohlke2023wireshape_CNC,Goshtasbi_2025_WIld_CNC} now enable the fabrication of highly customized chamber geometries from user-defined digital models. Combining CNC heat sealing with the symmetrical-chamber design and bottom constraint layer of our prior glove~\cite{Rui_2025_customized_glove}, we developed dual-action actuators for the soft robotic glove.

\begin{figure}[htbp]
    \centering
    \includegraphics[width=\columnwidth]{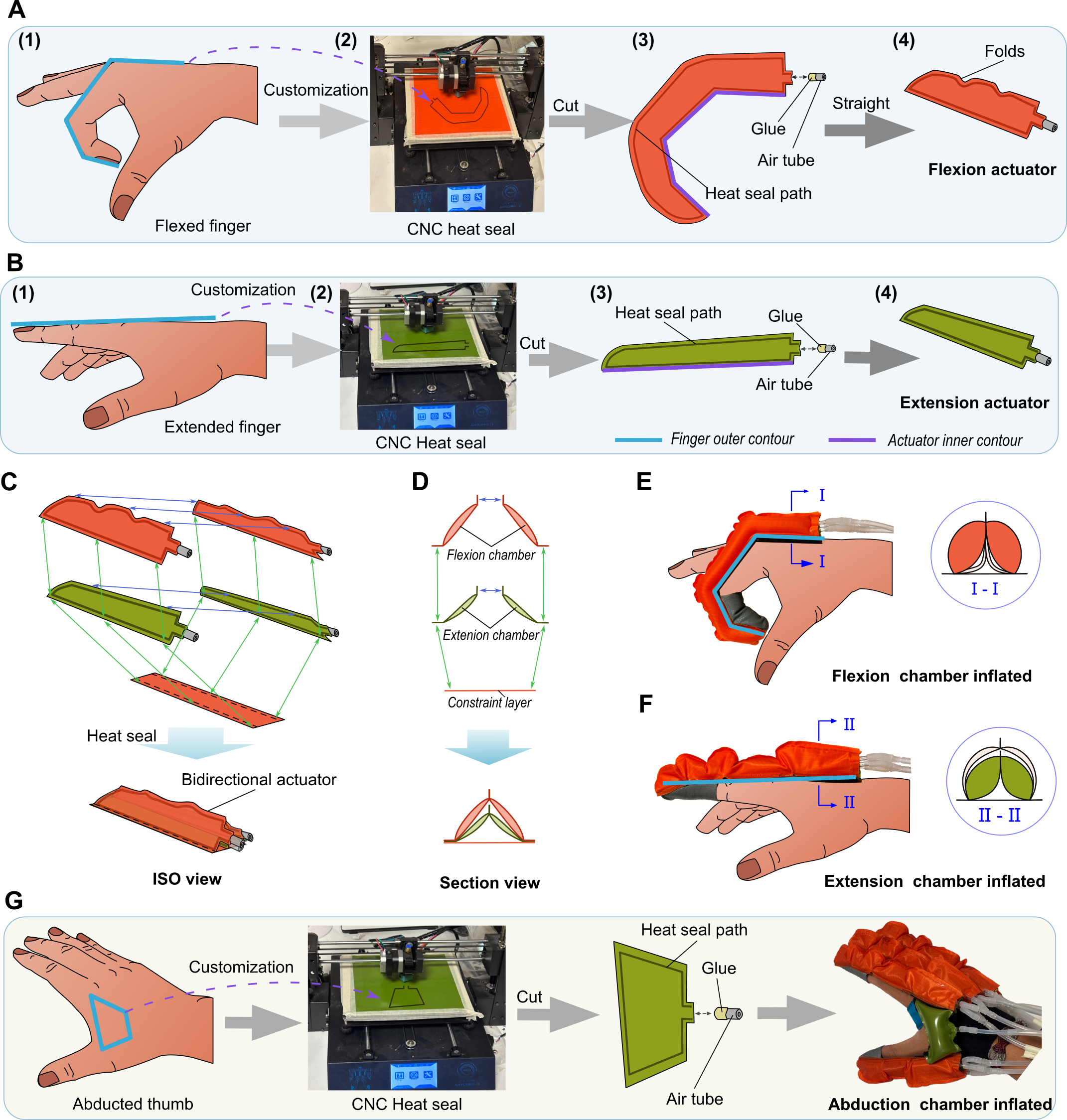}
    \caption{\textbf{Actuator design and fabrication.}
    \textbf{(A)} and \textbf{(B)} Customized flexion and extension actuators fabricated via CNC heat sealing.
    \textbf{(C)} and \textbf{(D)} dual-action actuator assembly by heat sealing flexion chambers, extension chambers, and a bottom constraint layer.
    \textbf{(E)} and \textbf{(F)} Actuator deflection toward the target flexion and extension postures upon selective chamber inflation.
    \textbf{(G)} Thumb abduction actuator fabrication process.}
    \label{Fig-M-Fabrication}
\end{figure}

Actuator fabrication begins by measuring the target joint angles at the desired flexed and extended postures directly from the patient's hand. These angles define the inflated geometry of each actuator segment. Prior work has established a near-linear relationship between the deflated actuator geometry and the inflated bending angle~\cite{andrade2023fabric_Star_Griper,chen2026_GPAs}, from which the required fabrication geometry---and thus the CNC sealing path---can be derived directly, as shown in Fig.~\ref{Fig-M-Fabrication}A and B. To assemble a dual-action actuator, two flexion chambers, two extension chambers, and a bottom constraint layer are heat sealed together (Fig.~\ref{Fig-M-Fabrication}C, D). Selective inflation of the respective chambers drives the actuator toward the target flexion or extension posture, providing individualized assistance in either direction (Fig.~\ref{Fig-M-Fabrication}E, F). The thumb abduction actuator is produced using the same fabrication procedure (Fig.~\ref{Fig-M-Fabrication}G). This finger-level process can be applied independently to each digit, allowing the glove configuration to be tailored to an individual's hand geometry.

\subsection*{Actuator Characterization}

\begin{figure}[htbp]
    \centering
    \includegraphics[width=\columnwidth]{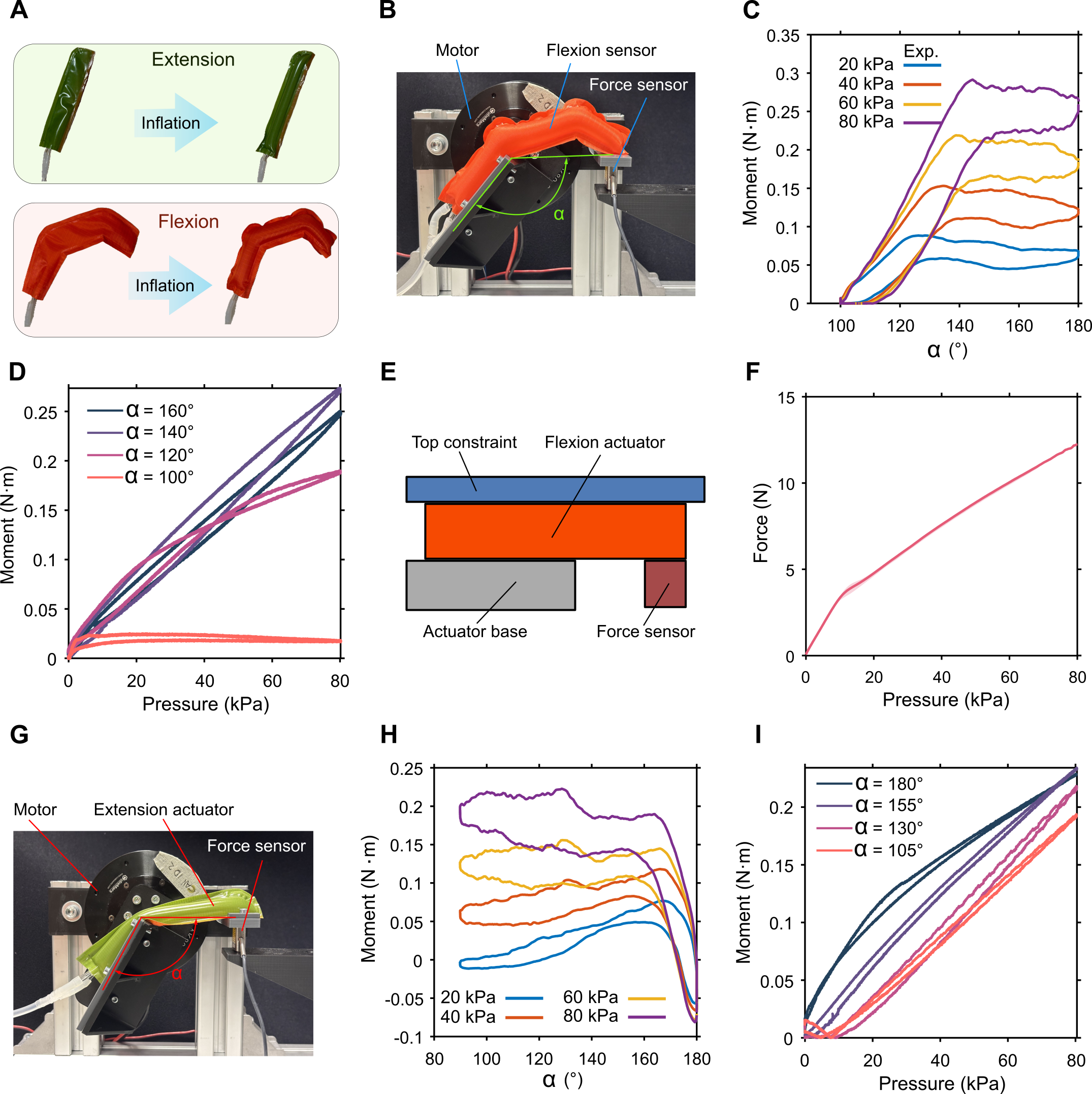}
    \caption{\textbf{Thumb actuator characterization.}
    \textbf{(A)} Extension and flexion actuator operating configurations.
    \textbf{(B)} Moment characterization experimental setup.
    \textbf{(C)} and \textbf{(D)} Effects of fold angle and inflation pressure on flexion bending moment.
    \textbf{(E)} and \textbf{(F)} Fingertip blocking force measurement setup and results.
    \textbf{(G)} Extension actuator moment characterization setup.
    \textbf{(H)} and \textbf{(I)} Effects of fold angle and inflation pressure on extension bending moment.}
    \label{Fig-M-ActuatorTest}
\end{figure}

We systematically characterized the force and bending moment output of the proposed actuators. The thumb actuator was selected as a representative case, with flexion and extension components evaluated separately (Fig.~\ref{Fig-M-ActuatorTest}A). Bending moment was measured as a function of fold angle and inflation pressure using a rotational servo motor and load cell (Fig.~\ref{Fig-M-ActuatorTest}B). Results showed that moment increased monotonically with both fold angle and inflation pressure, reaching a torque of 0.27~N$\cdot$m at 80~kPa for the flexion actuator (Fig.~\ref{Fig-M-ActuatorTest}C, D). Fingertip blocking force for the flexion actuator reached 12.2~N at 80~kPa (Fig.~\ref{Fig-M-ActuatorTest}E, F), comparable to previously reported fabric-based designs~\cite{Lai2023_Honey_Comb_glove,lim2023_Bidirectional_Fabric}. Characterization of the extension actuator under comparable conditions yielded a peak bending moment of 0.23~N$\cdot$m at 80~kPa (Fig.~\ref{Fig-M-ActuatorTest}G, H, I), with the combined five-finger extension moment exceeding 1~N$\cdot$m at that pressure. The slightly lower extension moment relative to flexion is due to the smaller actuator used. Together, these results confirm that the dual-action actuators provide sufficient mechanical output for both assistive and rehabilitative hand tasks.

\subsection*{Glove Grasping Performance}

\begin{figure}[htbp]
    \centering
    \includegraphics[width=\columnwidth]{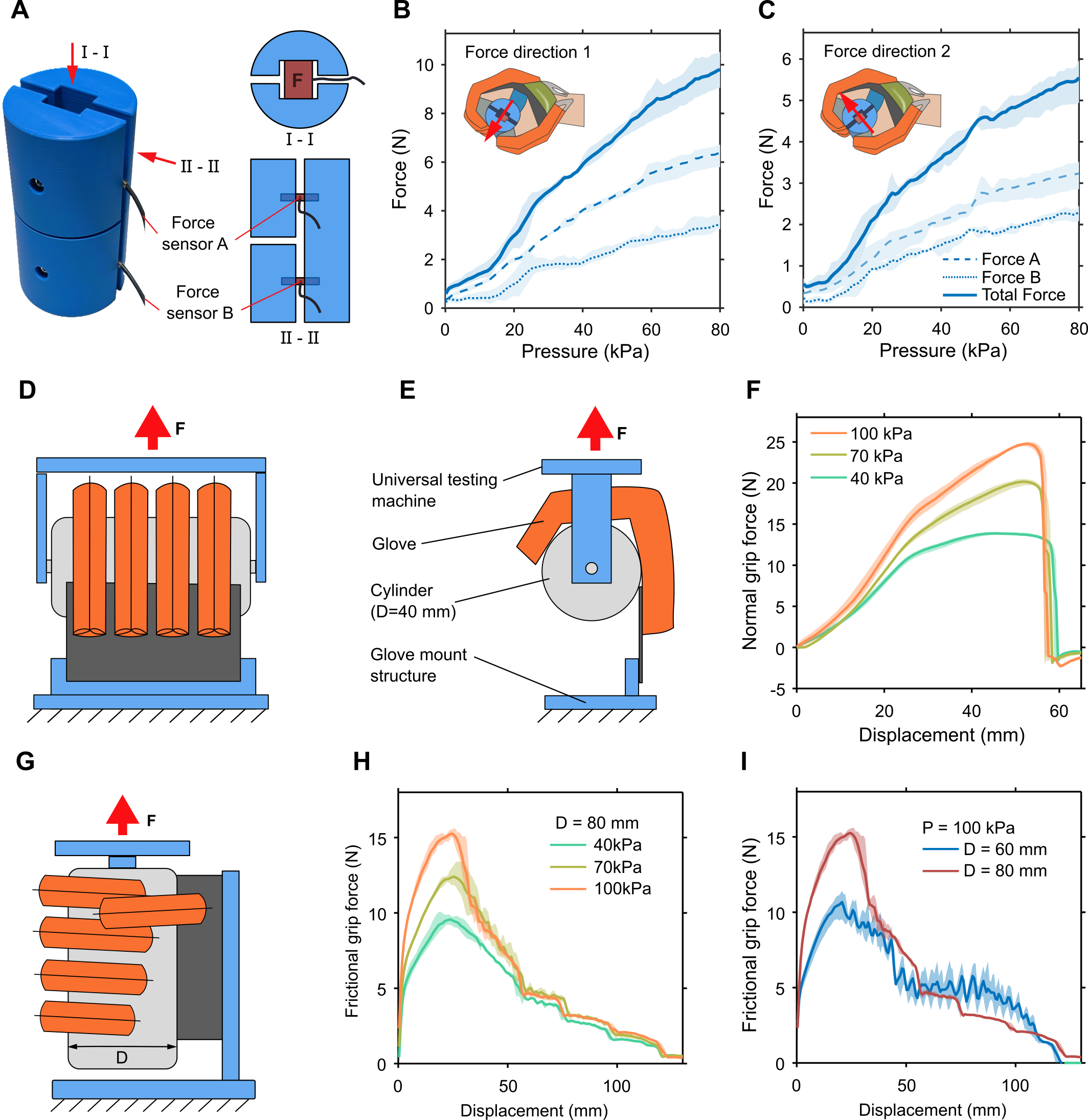}
    \caption{\textbf{Glove grasping performance characterization.}
    \textbf{(A)} Directional grasp force measurement setup.
    \textbf{(B)} and \textbf{(C)} Directional grasp force at different inflation pressures.
    \textbf{(D)} and \textbf{(E)} Normal grasp force measurement configuration.
    \textbf{(F)} Normal grasp force as a function of inflation pressure.
    \textbf{(G)} Frictional grasp force measurement setup.
    \textbf{(H)} and \textbf{(I)} Effects of inflation pressure and cylinder diameter on frictional grasp force.}
    \label{Fig-M-Glovetest}
\end{figure}

Grasp force is a critical performance metric for rehabilitation gloves, as it directly determines the range of ADL tasks the device can support. Force output was measured using two sensors positioned as illustrated in Supplementary Figure~1A. Grasp force exhibited directional variation (Fig.~\ref{Fig-M-Glovetest}B, C): at 80~kPa, the total pinch output reached 9.8~N, with the sensor proximate to the thumb consistently recording higher forces, consistent with the dominant role of the thumb in opposition-based grasping.

Normal grasping force, evaluated under the configuration shown in Supplementary Figure~1B, reached a maximum of 24.8~N at 100~kPa, demonstrating the glove's capacity to support power grasp tasks. This is comparable to previously reported fabric-based soft gloves~\cite{lim2023_Bidirectional_Fabric} and sufficient to assist with ADLs such as gripping bottles and cylindrical objects~\cite{2010p_roperties_of_Objects}.

Frictional grasping force, which governs the ability to hold objects against gravitational or resistive loads, was further characterized using the configuration shown in Supplementary Figure~1C. The effects of inflation pressure and cylinder diameter on frictional output were systematically investigated (Fig.~\ref{Fig-M-Glovetest}H, I). Frictional force increased with inflation pressure across all cylinder diameters tested, and larger diameters generally yielded higher friction due to increased contact area between the glove and the object surface. These results suggest that the glove is particularly well suited for grasping cylindrical objects commonly encountered in ADL-based rehabilitation protocols~\cite{2010p_roperties_of_Objects}.

Taken together, the characterization results demonstrate that the proposed glove provides mechanically sufficient and functionally relevant grasping performance across multiple grasp types, supporting its suitability for structured hand rehabilitation.

\subsection*{Preliminary study with healthy subjects}

\begin{figure}[htbp]
    \centering
    \includegraphics[width=\columnwidth]{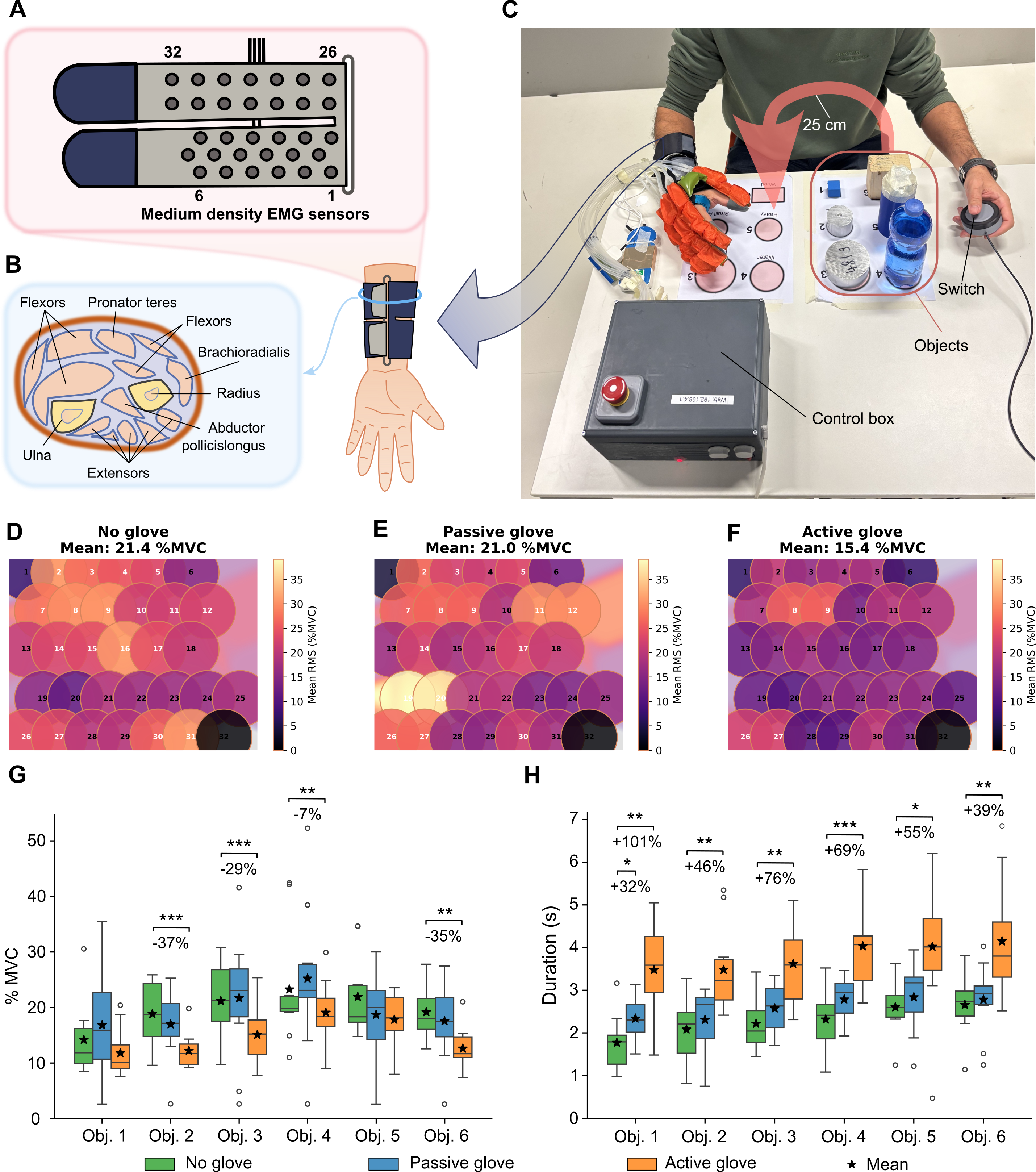}
    \caption{\textbf{Protocol and results of the preliminary healthy subjects study.}
    \textbf{(A)--(C)} Participants performing six transport tasks with the sEMG sleeve under three glove conditions (no glove, passive, and active).
    \textbf{(D)--(F)} Averaged sEMG amplitude across 31 active channels for Object~2, illustrating differences between conditions.
    \textbf{(G)--(H)} Mean muscle activity and task completion time across six objects under the three conditions.}
    \label{Fig-M-Healthy-Subjects}
\end{figure}

To evaluate the glove's capacity to reduce muscle activity during object manipulation, we conducted a preliminary study with ten healthy subjects (demographics in Supplementary Table~1). Participants transported six representative objects of varying dimensions and mass (Supplementary Figure~2A and Supplementary Table~2) while wearing a Medium-density surface electromyography (sEMG) sleeve with 32 channels to capture the forearm muscle activation throughout the tasks (Fig.~\ref{Fig-M-Healthy-Subjects}A--C; Supplementary Figures~2B and 2C). Three conditions were evaluated: no glove (bare hand), passive glove (worn but unactuated), and active glove (60~kPa applied to both flexion and extension channels for all participants).

Compared to the no-glove condition, active glove use reduced muscle activity across channels, whereas the passive glove condition did not significantly increase muscular effort, indicating that the device imposes minimal mechanical impedance when unactuated (Fig.~\ref{Fig-M-Healthy-Subjects}D--F). The degree of assistance varied across objects (Supplementary Figures~3--8). Statistically significant reductions in muscle activity were observed for Objects~2, 3, 5, and 6, with mean reductions of 37\% ($p = 0.001$), 29\% ($p = 0.001$), 7\% ($p = 0.007$), and 35\% ($p = 0.005$), respectively. For Object~4, the reduction approached but did not reach the significance threshold ($p = 0.053$). The button-switch actuation interface increased task completion time by 101\% for Object~1 ($p = 0.005$), 46\% for Object~2 ($p = 0.003$), 76\% for Object~3 ($p = 0.005$), 69\% for Object~4 ($p = 0.001$), 55\% for Object~5 ($p = 0.014$), and 39\% for Object~6 ($p = 0.001$), relative to the no-glove condition (all one-tailed Wilcoxon signed-rank tests). This additional time reflects the manual button-switch interface and the finite inflation and deflation latency of the pneumatic system rather than mechanical resistance from the glove, consistent with the absence of a significant time or muscle-activity increase in the passive condition. Despite this overhead, the consistent reductions in muscle activity for larger and heavier objects confirm that the glove provides effective mechanical assistance during functional manipulation tasks (Fig.~\ref{Fig-M-Healthy-Subjects}G--H; Supplementary Figure~2E).

\subsection*{Pilot Feasibility Study with Cervical SCI Patients}

Given that individuals with cervical SCI exhibit different muscle strength and coordination patterns compared to healthy subjects, and that the glove's primary target population is neurologically impaired users, we conducted a pilot feasibility study with three individuals with cervical SCI (2 male, 1 female; injury levels C5--C7; AIS grades A--D; time since injury 2--24 months), as shown in Fig.~\ref{Fig-M-Clinical-Results}A. Detailed participant demographics are provided in Supplementary Table~3.

Each participant completed a personalized subset of seven tasks (Fig.~\ref{Fig-M-Clinical-Results}B) selected according to their residual motor capacity: T1---box and block test, T2---water pouring test (GRASSP sub-task), T3---jar opening test (GRASSP sub-task), T4---nine-hole peg test, and three non-standardized object-transport tasks: T5---spherical grasp, T6---cylindrical grasp, and T7---lateral grasp.

For Task~1, Subjects~1 and 2 transferred fewer blocks per trial relative to the unassisted condition, likely attributable to the additional actuation time required per grasp cycle; in contrast, Subject~3 achieved a higher block count with glove assistance (Fig.~\ref{Fig-M-Clinical-Results}C). Notably, all three participants exhibited fewer object drops with the glove than in the unassisted condition, indicating improved grasp stability across the cohort.

For Task~2 (water pouring), Subject~1 was unable to initiate the task without the glove but could partially complete it with assistance, representing a clinically meaningful functional gain. Subject~2 could complete the task without the glove only through tenodesis grasp; with glove assistance, the same task was accomplished without this compensation, yielding a higher functional score. For Tasks~3 and 4, a modest performance improvement was observed for Subject~2 with glove assistance, while no notable difference was recorded for the other participants.

For Tasks~5 and 6, task completion time increased marginally with glove use; however, glove assistance enabled Subject~3 to eliminate tenodesis grasp entirely, achieving more anatomically consistent movement patterns. For Task~7, Subject~3 was unable to perform the task under either condition, and no meaningful difference was observed for Subject~2.

Although all participants required additional time for actuation, the overall results indicate that glove assistance enhanced hand function in several respects: promoting more natural grasp patterns, reducing reliance on tenodesis grasp, and decreasing the incidence of object drops (Supplementary Figure~9). These findings support the feasibility of the proposed approach in a clinical cervical SCI population.

\begin{figure}[htbp]
    \centering
    \includegraphics[width=\columnwidth]{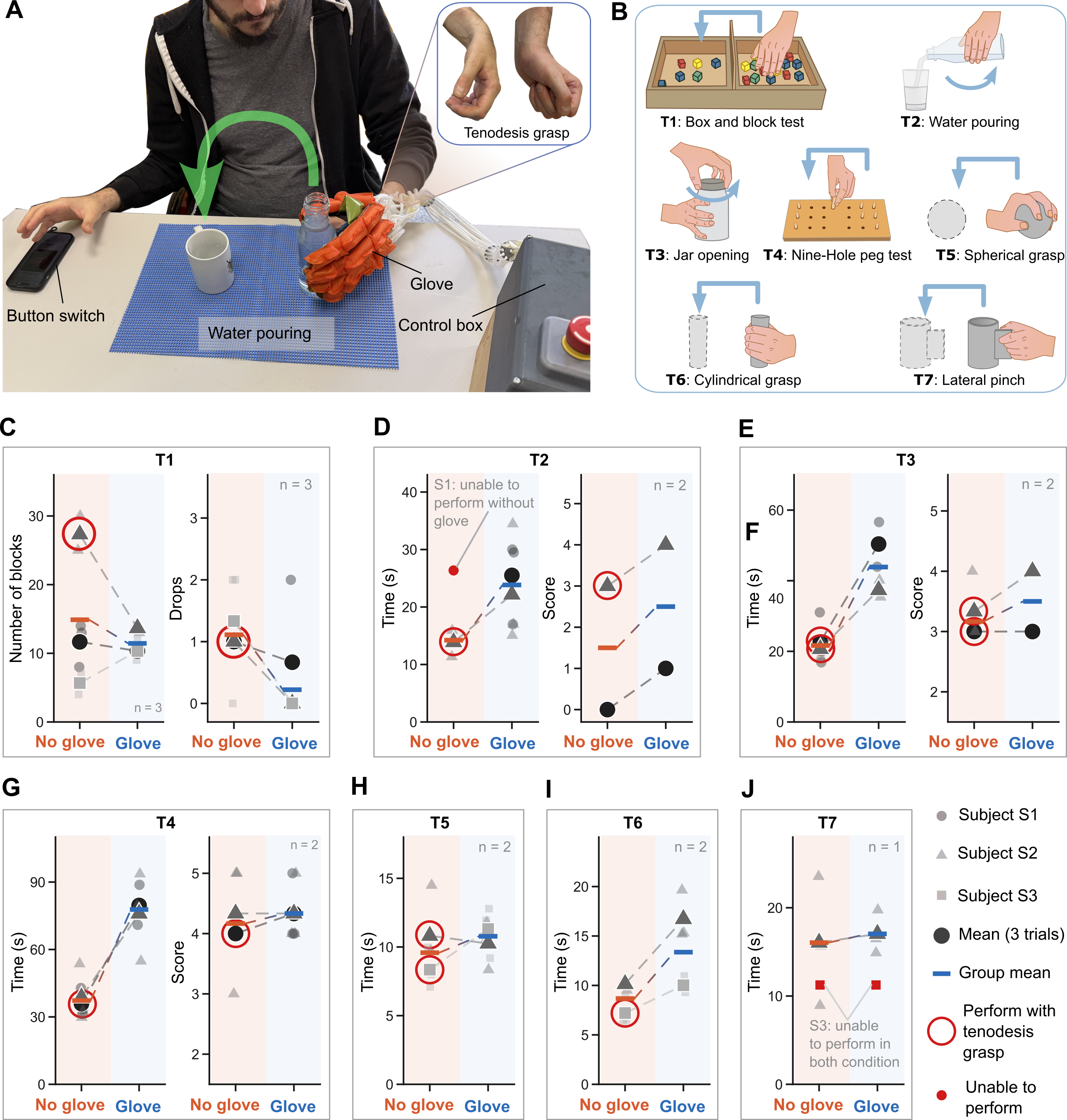}
    \caption{\textbf{Pilot clinical feasibility study.}
    \textbf{(A)} Experimental setup for clinical testing.
    \textbf{(B)} Overview of the seven evaluation tasks.
    \textbf{(C)--(J)} Task-level results for three individuals with cervical SCI under assisted and unassisted conditions.}
    \label{Fig-M-Clinical-Results}
\end{figure}

\section*{Discussion}

We have developed a dual-action fabric-based soft robotic glove that addresses key limitations of current hand rehabilitation devices through geometric customizability and ergonomic design.

\textbf{Symmetrical-chamber architecture and ergonomic conformity.}
The symmetrical-chamber actuator offers a practical alternative to conventional single-chamber pneumatic actuators. Its two symmetric chambers, shaped by CNC heat sealing, serve a dual role: their symmetry yields a concave contact surface that conforms to the convex finger dorsum and is intended to distribute contact pressure more evenly, while their CNC-defined profile, mapped to the inflated bending angle by the near-linear relationship from our prior work~\cite{chen2026_GPAs}, can be tailored to a wearer's joint angles. Whereas our prior symmetrical-chamber glove provided flexion assistance only~\cite{Rui_2025_customized_glove}, here we extend the design to a dual-action glove adding extension and thumb abduction, a degree of freedom often absent from existing designs, while retaining the compliance and low weight of fabric-based actuators. The same profile customization can also target different postures, such as the distinct finger configurations of a power grasp versus a pinch, pointing to future potential for task-oriented training.

\textbf{Mechanical performance and assistive effectiveness.}
The characterization confirms that the actuators and glove generate forces adequate for ADL-relevant tasks, comparable to previously reported fabric-based designs~\cite{Lai2023_Honey_Comb_glove,lim2023_Bidirectional_Fabric}. The passive glove produced no significant increase in muscular effort over the bare hand, indicating negligible mechanical impedance when unactuated, which suits continuous wear across both active and passive phases of a session. The consistent EMG reductions for larger and heavier objects further suggest that the assistive benefit scales with task demand.

\textbf{Positioning relative to existing soft rehabilitation gloves.}
We compared the proposed glove against representative soft rehabilitation gloves spanning pneumatic, cable-driven, and shape-memory-alloy actuation (Supplementary Tables~4 and~5). Several prior fabric-based gloves report fingertip or grasping forces comparable to, or higher than, those measured here. The design objective of a wearable rehabilitation glove, however, differs from that of a standalone gripper optimized for maximal output: the target is force sufficient for activities of daily living while remaining safe and comfortable on the hand, rather than maximal. The fingertip force (12.2~N at 80~kPa), grasp force (9.8~N at 80~kPa), and normal grasping force (24.8~N at 100~kPa) reported here fall within the range required for common ADL objects~\cite{2010p_roperties_of_Objects}. What distinguishes the present device is therefore not any single metric but the combination, within a 90~g all-fabric package, of ergonomic concave contact, geometric customizability, dual-action assistance, and thumb abduction; to our knowledge, no prior design in this comparison offers all four.

\textbf{Functional completeness and clinical relevance.}
Many existing soft gloves provide flexion assistance only, limiting their utility for users who also need active finger extension for volitional object release in cervical SCI~\cite{lim2023_Bidirectional_Fabric,cappello2018_Assisting_silicone_glove}. By adding active extension and thumb abduction in a single lightweight device, the proposed glove supports a more complete set of functional movements. In the pilot SCI study, glove assistance was associated with new task initiation, reduced reliance on tenodesis strategies, and fewer object drops. These preliminary findings suggest potential clinical value, though the study did not isolate the contribution of any individual design element and the small sample precludes definitive conclusions.

\textbf{Limitations and future directions.}
Several limitations remain. First, actuators are fabricated individually from manual hand measurements, and customization was not applied systematically: the present devices used representative hand dimensions, with dimensional adaptation demonstrated in only one participant; automated measurement and a controlled comparison of customized versus standardized configurations are needed. Second, both studies are small and demographically narrow (ten healthy males, and three individuals with cervical SCI over a few sessions), so the findings are preliminary; larger, more diverse, and longitudinal trials, including individuals with stroke, are required to establish efficacy. Third, the study used only two whole-hand gestures under simultaneous actuation; although the control box supports per-finger actuation, its potential for task-specific grasps such as pinch, and for task-oriented training, was not evaluated~\cite{li2024Task-oriented,lee2024Task-oriented}. Fourth, the button-switch interface increased task completion time; more natural intent detection such as surface EMG or wrist motion sensing could reduce this overhead. Fifth, the thumb abduction actuator provides only a single direction; active bidirectional abduction could further extend the achievable thumb workspace. Finally, further miniaturization of the control box and assessment of long-term actuator durability remain for future work.

\textbf{Conclusion.}
In summary, the proposed glove demonstrates that a customizable, ergonomically informed design can improve both comfort and functional performance in hand rehabilitation. Realized through CNC heat sealing, the symmetrical-chamber actuators are designed to conform to the finger dorsum and to provide active flexion, extension, and thumb abduction across all five digits, with mechanical output adequate for ADL-relevant tasks. Evaluation in healthy subjects and individuals with cervical SCI provides encouraging preliminary evidence of functional utility; further controlled clinical investigation is needed to establish therapeutic efficacy and support translation to practice.

\section*{Materials and Methods}

\subsection*{Actuator and Glove Materials}

All actuator chambers were fabricated from thermoplastic polyurethane (TPU)-coated nylon fabric (Adventure Expert, Slovenia). Two fabric variants were used depending on their functional role: 40D ripstop nylon with single-sided TPU coating was used for the flexion-side outer layers (depicted in orange in Fig.~\ref{Fig-M-Fabrication}), providing higher tensile stiffness to resist outward expansion during flexion inflation; 20D nylon ripstop with double-sided TPU coating was used for the extension-side layers and the bottom constraint layer (depicted in green). The double-sided TPU coating of the 20D fabric ensures reliable thermal bonding between adjacent layers during assembly. TPU pneumatic tubing (outer diameter 4~mm, inner diameter 2.5~mm) connected the actuator chambers to the solenoid valves. Tubes were inserted through a small incision at the proximal end of each chamber and secured with cyanoacrylate adhesive. Velcro hook-and-loop fasteners were used for all actuator-to-base and strap attachment points. Fingertip sleeves were adapted from commercially available gloves and heat sealed to the distal end of each dual-action actuator. For the clinical SCI study, all fingertip sleeves were additionally coated with a thin layer of platinum-cure silicone rubber (Smooth-On Mold Max 40, Shore A 40) applied by brush and allowed to cure at room temperature for 24~h, to increase friction between the glove surface and grasped objects.

\subsection*{CNC Heat Sealing Fabrication and Glove Assembly}

A commercial FDM 3D printer (Anycubic Mega, Anycubic, China) was modified into a CNC thermal sealing platform by replacing the extruder assembly with a custom resistive heating element. Sealing path coordinates were derived from 3D geometric models of the target actuator contours and exported as G-code for the CNC platform; all corresponding STL files are provided in the Supplementary Material. Following CNC heat sealing, chamber outlines were manually trimmed along the outer sealed boundary, leaving a 4~mm bonded margin to prevent delamination under inflation pressure.

Flexion and extension chambers were paired for each finger and heat sealed together with a bottom constraint layer to form a dual-action actuator. Assembly-layer heat sealing was performed manually using a flat-tip handheld heat sealer at 200 $^\circ$C. Each completed dual-action actuator was fitted with a fingertip sleeve at the distal end and a Velcro patch at the proximal end for attachment to the fabric hand base. Five dual-action finger actuators were arranged on the fabric base. The thumb abduction actuator, fabricated following the same procedure, was positioned between the thumb and index finger and attached via Velcro. Three adjustable Velcro straps at the dorsal hand, wrist, and forearm provided secure anchoring of the glove to the user's hand.

\subsection*{Pneumatic Control System}

A portable silent piston air compressor (FIAC, Italy) provided regulated compressed air at up to 800~kPa for all characterization experiments. Proportional pressure control valves (ITV0010, SMC, Japan) regulated outlet pressure to each actuator channel. An ESP32 microcontroller (Espressif Systems) issued pressure commands via the analog output interface of the ITV0010 valves and received pressure feedback from the valves' built-in sensors.

The wearable control box housed the ESP32 microcontroller, 12 N-channel MOSFET gate drivers, 12 miniature solenoid on/off valves (operating voltage 12~V DC), a lithium polymer battery pack, and a step-down voltage regulation board (Supplementary Figure~2D). The 12 solenoid channels provided independent binary switching (inflate or deflate) for each actuator port: two channels per dual-action finger actuator (one flexion, one extension) for five fingers, plus one channel for the thumb abduction actuator. The microcontroller communicated with a smartphone via a Wi-Fi access point hosted on the ESP32 through a browser-based graphical interface. Predefined gesture programs (full flexion, full extension, index pointing, and lateral pinch) were implemented as sequential valve switching commands; in the present study, only the full flexion and full extension programs were used during experimental sessions.

\subsection*{Actuator Mechanical Characterization}

Bending moment was characterized using a servo motor (CubeMars AK80-8, T-Motor) mounted rigidly to a steel frame. The actuator was fixed at its proximal end to a clamp attached to the motor shaft. As the motor rotated the actuator at a constant angular velocity of 3$^\circ$/s through the target range of fold angles, a single-axis load cell (LSB205, Futek, USA) measured the reaction force at a fixed moment arm, from which bending moment was computed. Inflation pressure was held constant during each sweep by the ITV0010 valve. The load cell signal was sampled at 80~Hz and processed with a 20-point moving average filter. Fingertip blocking force was measured with the actuator fixed in the straight (unbent) configuration, with a load cell positioned at the fingertip contact point.

\subsection*{Glove Grasping Force Characterization}

For directional grasp force measurement, a custom cylinder (diameter 40~mm) was instrumented with two LSB205 load cells (Futek, USA) positioned on opposite sides to measure the force components exerted by each side of the glove independently, thereby avoiding internal reaction force cancellation that would occur with a single centrally mounted sensor. One subject was instructed to wear the glove and maintain a fully relaxed hand posture throughout the measurement. Inflation pressure was increased incrementally and the corresponding force output was recorded at each step. Three repetitions were conducted and results are reported as mean $\pm$ standard deviation.

Normal and frictional grasping forces were evaluated using a universal materials testing machine (FMT-313, PPT GmbH, Germany) equipped with a 50~N load cell. For normal force measurement, the glove was donned on a rigid hand frame secured to the lower fixture of the testing machine. A rotatable rigid cylinder was placed at a standardized starting position between the glove-covered fingers, and the upper fixture displaced the cylinder vertically upward at 200~mm/min while the compressive grasp force was recorded as a function of displacement. For frictional force measurement, the cylinder was displaced axially at the same traverse speed while the glove maintained a fixed inflation pressure of 100~kPa, with the displacement direction opposing the frictional resistance generated by the grasp. Two cylinder diameters (60~mm and 80~mm) were tested to characterize the effect of object geometry on frictional output. Three repetitions were conducted for each condition and averaged for analysis.

\subsection*{Healthy Subjects Study}

\subsubsection*{Participants}

A preliminary user study was conducted with 10 healthy male subjects (height: $180.4 \pm 7.1$~cm; weight: $80.0 \pm 9.6$~kg; age: $29.9 \pm 3.7$~years; hand length: $20.0 \pm 0.4$~cm; full demographic data in Supplementary Table~1). All participants reported no history of upper-limb injury or neurological disease. All experimental tasks were performed on the right hand. The study protocol was approved by the Research Ethics Committee of Scuola Superiore Sant'Anna, and all participants provided written informed consent prior to participation.

\subsubsection*{Experimental Protocol}

Six objects of varying dimensions and mass, representative of common household items encountered in ADL-based rehabilitation, were prepared (Supplementary Table~2). Participants were instructed to pick up each object with minimal effort, transport it horizontally 25~cm, and set it down, proceeding sequentially from Object~1 to Object~6. Three glove conditions were evaluated: no glove (bare hand), passive glove (worn but unactuated, to assess added mechanical impedance), and active glove (60~kPa applied simultaneously to both the flexion and extension channels). Three repetitions were completed per object per condition. Condition order was randomized across participants to reduce order and learning effects. Actuation was controlled by the participant via a handheld button switch held in the contralateral hand, enabling manual toggling between the full extension and full flexion gestures.

\subsubsection*{sEMG Recording and Processing}

Muscle activity was recorded using a customized medium-density sEMG sleeve with 32 independent channels (31 active, 1 non-functional), adapted from our prior work~\cite{mendez2025sEMG} (Supplementary Figures~2B and 2C). The sleeve is easy to don and doff and provides coverage of the relevant forearm region. Data were sampled at 1000~Hz and preprocessed using a 50~Hz notch filter ($Q = 30$) and a 15--450~Hz fifth-order Butterworth band-pass filter. Muscle activation was quantified using RMS features computed with 0.25~s windows (0.15~s shift), and amplitudes were normalized to each participant's maximum voluntary contraction (MVC) and expressed as \%MVC~\cite{Rui_2025_customized_glove,Rui_2025_LPPAMs}. Using the recorded time labels, pick-and-place epochs were extracted for each of the six objects. The RMS amplitude of each epoch was computed per channel, object, and condition, yielding 30 values per combination (10 subjects $\times$ 3 repetitions), with occasional reduction to 29 due to data loss.

Statistical analysis employed the one-tailed Wilcoxon signed-rank test for both task completion time and normalized muscle activation (\%MVC), following confirmation of non-normality in several conditions via the Shapiro--Wilk test. Directional hypotheses were specified a priori as follows: the active glove was expected to reduce muscle activation (Active $<$ No glove) and increase task completion time (Active $>$ No glove); the passive glove was expected to increase both muscle activation (Passive $>$ No glove) and task completion time (Passive $>$ No glove) relative to the no-glove condition. The absence of significant differences in the passive glove comparisons was interpreted as evidence of minimal mechanical impedance introduced by the passive structure. Significance levels are denoted as $^{***}$: $p < 0.005$; $^{**}$: $p < 0.01$; $^{*}$: $p < 0.05$.

\subsection*{Pilot Feasibility Study in Cervical SCI}

\subsubsection*{Participants}

Two individuals with chronic and one with acute cervical SCI were recruited from Careggi Hospital (Unit\`{a} Spinale Unipolare), Florence, Italy. Injury levels ranged from C4 to C7, with AIS grades from A to D. Time since injury ranged from 2 to 24 months. All participants presented with clinically confirmed upper-limb motor impairment affecting hand function. Detailed demographic and clinical characteristics are summarized in Supplementary Table~3. Given the limited sample size inherent to a pilot study, findings should be interpreted as preliminary evidence of device feasibility rather than definitive clinical efficacy.

All ethical and experimental procedures were approved by the Ethics Committee of Careggi University Hospital (Florence, Italy) in compliance with the Declaration of Helsinki and applicable Good Clinical Practice guidelines. All participants provided written informed consent prior to enrolment.

\subsubsection*{Experimental Protocol}

Seven functional tasks were administered to assess the effect of glove assistance on upper-limb performance (details of task see Supplementary notes). The first four tasks followed established standardized protocols, while the remaining three were purpose-designed for this study. Task~1 was the box and block test~\cite{desrosiers1994Box_and_block}: participants transferred as many blocks as possible across a central partition within 1~minute; the number of successful transfers and accidental drops were recorded to quantify gross hand dexterity. Tasks~2 and 3 comprised sub-tasks from the GRASSP clinical assessment protocol~\cite{kalsi2009GRASSP1,kalsi2012Grasp2}: a water pouring test and a jar opening test, respectively. Task~4 was the nine-hole peg test, in which participants transferred nine pegs into a pegboard; total completion time was recorded as a measure of fine manual dexterity. For Tasks~5--7, three lightweight objects were custom 3D-printed to elicit distinct grasp patterns: a sphere (34~g, $\phi$70~mm) for spherical grasp, a cylinder (22~g, $\phi$30~$\times$~130~mm) for cylindrical grasp, and a handled cup (58~g, 5~mm wall thickness) for lateral grasp. Participants were instructed to transport each object to a designated target location. These three tasks were inspired by~\cite{light2002Task_references} and are compatible with the glove's actuation design.

Each participant completed three repetitions of each assigned task under two conditions: with and without glove assistance. Participants operated the actuation system using a contralateral hand-held button switch. Task allocation was individualized according to each participant's residual motor capacity: Subject~1 completed Tasks~1--4; Subject~3 completed Tasks~1 and 5--7; Subject~2 completed all seven tasks. To accommodate Subject~3 (female), the index and middle finger actuators were replaced with smaller-sized counterparts, and active assistance was restricted to the thumb, index, and middle fingers. All fingertip sleeves were coated with a thin silicone layer (Smooth-On Mold Max 40) to enhance friction with grasped objects.

The glove was fitted to the more affected hand as determined by clinical assessment. Subject~1 was assisted on the right hand; Subjects~2 and 3 received assistance on the left hand. Actuation pressure was individually calibrated prior to the experimental session based on subjective comfort and perceived adequacy of assistance: Subject~1 at 70~kPa, Subject~2 at 50~kPa, and Subject~3 at 60~kPa. Pressure was adjusted iteratively during a brief familiarization protocol until the participant reported satisfactory assistive force without discomfort.

\subsubsection*{Performance Scoring}

For Tasks~2--4, task performance was quantified using a six-level ordinal scoring rubric adapted from the GRASSP protocol~\cite{kalsi2012Grasp2}, with a maximum allowable duration of 2~min per task:

\begin{itemize}
    \item[\textbf{0}] --- The task cannot be initiated.
    \item[\textbf{1}] --- Less than 50\% of the task is completed; the expected grasp pattern is not achieved.
    \item[\textbf{2}] --- 50\% or more of the task is completed; the expected grasp pattern is not achieved.
    \item[\textbf{3}] --- The task is completed using tenodesis or an alternative compensatory grasp pattern.
    \item[\textbf{4}] --- The task is completed using the expected grasp pattern with difficulty (slow or non-fluid movement).
    \item[\textbf{5}] --- The task is completed fluently using the expected grasp pattern.
\end{itemize}

\noindent The 50\% completion threshold for Task~2 is defined as the onset of liquid pouring; for Task~4, it is defined as the moment the participant positions a peg at the pegboard insertion point.
%%%%%%%%%%%%%%%%%%%%%%%%%%%%%%%%%%%%%%%%%%%%%%%%%%%%%%%%%%%%%%%%%%%%%%%%%%%%%%%%%%%%%%%%%%%%%%%%%%%%%%%%%%%%%%%%%%%%%%%%%%%%%%%%%%%%%%%%%%%%
\section*{Declaration statements}

\subsection*{Data Availability}

All data are available in the main text or the supplementary materials. The control software, and experimental datasets supporting the conclusions of this article are available from the corresponding author upon reasonable request.

\subsection*{Acknowledgments}

This project was funded by MSCA-DN / Project 101073374 - ReWIRE. Views and opinions expressed are however those of the author(s) only and do not necessarily reflect those of the European Union or the European Research Executive Agency (REA). Neither the European Union nor the granting authority can be held responsible for them.

% \subsection*{General} 

\subsection*{Author Contributions} 

R.C.: Conceptualization, Methodology, Investigation, Formal analysis, Visualization, Writing – original draft, Validation.
F.I.S.: Investigation, Methodology, Visualization. 
D.C.: Methodology, Investigation, Writing – review and editing.
X.M.: Investigation, Validation.
E.L.: Methodology, Resources, Writing – review and editing.
G. R.: Methodology, Investigation, Data curation, Resources.
C.D.S: Methodology, Investigation, Data curation.
F. S.: Methodology, Investigation.
C.C.: Methodology.
V.M.: Resources.
D.L.: Methodology, Supervision.
G. D. P.: Supervision, Methodology, Resources.
S.M.: Supervision, Funding acquisition, Writing – review and editing.
A.F.: Supervision, Funding acquisition, Writing – review and editing.
All authors reviewed and approved the final manuscript.

\subsection*{Conflicts of Interest}
The authors declare that there is no conflict of interest regarding the publication of this article.

\printbibliography

\end{document}